\documentclass[11pt,a4paper]{article}
\usepackage[hyperref]{emnlp-ijcnlp-2019}
\usepackage{times}
\usepackage{latexsym}
\usepackage{amsmath}
\usepackage{url} 
\usepackage{graphicx}
\usepackage{booktabs}
\aclfinalcopy % Uncomment this line for the final submission

\title{Toward Understanding The Effect Of Loss function On Then Performance Of Knowledge Graph Embedding}

\author{Mojtaba Nayyeri \\
  University of Bonn, Bonn, Germany \\
  \texttt{nayyeri@cs.uni-bonn.de} \\\And
  Chengjin Xu \\
  University of Bonn, Bonn, Germany \\
  \texttt{xuc@cs.uni-bonn.de} \\\AND
  Yadollah Yaghoobzadeh \\
  Microsoft Research Montŕeal \\
  \texttt{yayaghoo@microsoft.com} \\\And
  Hamed Shariat Yazdi \\
  University of Bonn, Bonn, Germany \\
  \texttt{Shariat@cs.uni-bonn.de}\\\AND
   Jens Lehmann \\
  University of Bonn, Bonn, Germany \\
  Fraunhofer IAIS, Bonn, Germany \\
  \texttt{jens.lehmann@cs.uni-bonn.de}\\
}

\date{}

\begin{document}
\maketitle

\begin{abstract} 
Knowledge graphs (KGs) represent world's facts in structured forms. 
KG completion exploits the existing facts in a KG to discover new ones. Translation-based embedding model (TransE) is a prominent formulation to do KG completion. 
Despite the efficiency of TransE in memory and time, it suffers from several \textit{limitations} in encoding relation patterns such as  symmetric, reflexive etc. To resolve this problem, most of the attempts have circled around the revision of the \textit{score} function of TransE i.e., proposing a more complicated score function such as Trans(A, D, G, H, R, etc) to mitigate the limitations. 
In this paper, we tackle this problem from a different perspective. We show that existing theories corresponding to the limitations of TransE are inaccurate because they ignore the effect of \textit{loss} function. Accordingly, we pose theoretical investigations of the main \textit{limitations} of TransE in the light of \textit{loss} function. 
%rather than the score function. 
To the best of our knowledge, this has not been investigated so far comprehensively. We show that by a proper selection of the loss function for training the TransE model, the main limitations of the model are mitigated. This is explained by setting upper-bound for the scores of positive samples, showing the region of truth (i.e., the region that a triple is considered positive by the model).
Our theoretical proofs with experimental results fill the gap between the capability of translation-based class of embedding models and the loss function. The theories emphasise the importance of the selection of the loss functions for training the models. 
Our experimental evaluations on different loss functions used for training the models justify our theoretical proofs and confirm the importance of the loss functions on the performance.

\end{abstract}

\section{Introduction}
%\textbf{Topic1: Importance Of Knowledge in Human life}

Knowledge is considered as commonsense facts and other information accumulated from different sources. Throughout history, civilizations have evolved due to increase in the knowledge. With the passage of time, humans obtain many 
relations among different entities. Therefore, development of proper knowledge representation (KR) and management systems is essential. 

%\textbf{Topic2: Importance Of Knowledge representation, focus on KG:}

The aim of KR is to study how the beliefs can be represented in an explicit, symbolic notation proper to automated reasoning. 
%KR plays a central role in AI. 
%New trends in AI ask for a new paradigm for KR and reasoning at the scale of semantic web. 
Knowledge Graph (KG) is a new direction for KR.
KGs are usually represented as a set of triples ($h,r,t$) where $h,t$ are entities and $r$ is a relation, e.g.\ $(iphone, hyponym, smart phone)$. 
Entities and relations are nodes and edges in the graph, respectively. 

%\textbf{Topic3: Problem of KGs (incomplete) and solution with KGE for completion:}

KGs are inherently incomplete, making prediction of missing links always relevant. Among different approaches used for KG completion, KG Embedding (KGE) has recently received growing attentions.  
% KGE assumes that there are global features representing KG. 
KGE embeds entities and relations as low dimensional vectors.  To measure the degree of plausibility of a triple,
a scoring function is defined over the embeddings.

%\textbf{Topic4: TransE as a KGE for KG completion:}

TransE, Translation-based Embedding model,
\cite{bordes2013translating} is one of the most widely used KGEs. The original assumption of TransE is to hold: 
$\textbf{h} + \textbf{r} = \textbf{t}$, 
for every positive triple ($h,r,t$)
where $\textbf{h},\textbf{r},\textbf{t} \in R^d$ are embedding vectors of head ($h$), relation ($r$) and tail ($t$) respectively. 

%\textbf{Topic5: limitation of TransE:}

TransE and its many variants like TransH \cite{transH}
and TransR \cite{transr}, underperform greatly compared to the current state-of-the-art embedding models due to the inherent limitations of their scoring functions. 

Recent work has the main limitations of Translation-based models. 
\cite{expressive} reveals that TransE cannot encode a relation pattern which is neither reflexive nor irreflexive. \cite{sun2019rotate} prove that TransE is incapable of encoding symmetric relation. \cite{transH} 
adds that TransE 
% can encode irreflexive and one-to-one relations while it 
cannot properly encode reflexive, one-to-many, many-to-one and many-to-many relations. 
%\newcite{transH} argues that to encode reflexive relations, the condition $\textbf{h} + \textbf{r} = \textbf{h}$ should be held, which means that $\textbf{r} = \textbf{0}$ and all entities will have the same vectors. 
%For one-to-many relations, TransE gives the same embedding vectors for many entities which are used as object (tail entity), which is undesired. 
%\begin{equation}
%    \begin{array}{lr}
%     \textbf{h} + \textbf{r} = \textbf{t}_1, &\\
%     \textbf{h} + \textbf{r} = \textbf{t}_2,\\
%     \textbf{h} + \textbf{r} = \textbf{t}_3,\\
%     ... \\
%     \textbf{h} + \textbf{r} = \textbf{t}_M,
%    \end{array}
%\end{equation}

%Therefore, $\textbf{t}_1 = ...=\textbf{t}_M$ and all entities $\textbf{t}_1, ...,\textbf{t}_M$ will have the same vector representations.

TransH, TransR and TransD \citep{transH,transr,transD2015} can handle the mentioned problems of TransE (i.e.\ one-to-many, many-to-one, many-to-many and reflexive) by projecting entities to relation space before applying translation. 
%Thus, projected vectors will be represented by  same vectors while original embedding vectors of entities will be unique. 
However, \cite{kazemi2018simple} investigate three additional limitations of TransE, FTransE \citep{ftanse2016}, STransE \citep{stranse2016}, TransH and TransR models: (i) if the models encode a reflexive relation $r$, they automatically encode symmetric, (ii) if the models encode a reflexive relation $r$, they automatically encode transitive and, (iii) if entity $e_1$ has relation $r$ with every entity in $\Delta \in \mathcal{E}$ and entity $e_2$ has relation $r$  with one of entities in $\Delta$, then $e_2$ must have the relation $r$ with every entity in $\Delta$. 

%\textbf{Topic6: Inaccuracy of limitation of transx models:}

The mentioned works have investigated these limitations by focusing on the capability of \textit{scoring} functions in encoding relation patterns. However, we prove that the selection of \textit{loss} function affects the boundary of score functions; consequently, the selection of loss functions significantly affects the limitations. Therefore, the above mentioned theories corresponding to the limitations of translation-based embedding models in encoding relation patterns are inaccurate. We pose new theories about the limitations of TransX(X=H,D,R, etc) models considering %both \textit{scoring} and 
the \textit{loss} functions. To the best of our knowledge, it is the first time that the effect of loss function is investigated to prove theories corresponding to the limitations of translation-based models. 

%In contrast to TransH, TransR projects entities to relation space by using a matrix provided for each relation. Using matrix for projection gives the model flexibility to provide different spaces for entities and relations. 
%\textbf{Topic6: summary of our contribution:}
In a nutshell, the key contributions of this paper is as follows: 
(i) We show that different loss functions enforce different upper-bounds and lower-bounds for the scores of positive and negative samples respectively. This implies that existing theories corresponding the limitation of TransX models are inaccurate because the effect of loss function is ignored. We introduce new theories accordingly and prove that the proper selection of loss functions mitigates the main limitations. 
(ii) We reformulate the existing loss functions and their optimization problems as an standard constrained optimization problem. This makes perfectly clear that how each of the loss functions affect on the boundary of triples scores and consequently ability of relation pattern encoding. 
(iii) using symmetric relation patterns, we obtain the optimal upper-bound of positive triples score to enable encoding of symmetric patterns.
(iv) We additionally investigate the theoretical capability of translation-based embedding model when translation is applied in complex space (TransComplEx). We show that TransComplEx is a more powerful embedding model with fewer theoretical limitations in encoding different relation patterns such as symmetric while it is efficient in memory and time. %Considering the effect of \textit{score} function on the limitations, we propose TransComplEx, a new variant of TransE that applies translation in complex space. TransComplEx is a more powerful embedding model with fewer theoretical limitations in encoding different relation patterns such as symmetric patterns. %we show the reason behind the positive effect of normalization of entities on the performance when a Margin Ranking Loss is used for training the TransX models. (ii) We show that the previous theories about limitations of translational embedding models are inaccurate, since they ignore the effect of loss function. We introduce theories accordingly and also using loss functions enforcing different upper- and lower-bounds for scores of positive and negative examples, respectively, mitigates the limitations.
%(iii) We derive formulae to encode some relation patterns including symmetric, equivalence, implication and inverse in our model.

%The rest of this paper is organized as follows: Section \ref{relatedwork} presents preliminaries and related works. Our model and theories are presented in the section \ref{OurModel}. The proposed model and the theories are experimentally evaluated in the section \ref{exper}. Finally the paper is concluded in the section \ref{conclusion}. 

\section{Related Works}\label{relatedwork}
%In this paper we extend the score function of TransE from real space to the complex space which is less restrictive in encoding relation patterns according to our theories. 
Most of the previous work have investigated the capability of translation-based class of embedding models considering solely the formulation of the score function. 
Accordingly, in this section, we review the score functions of TransE and some of its variants together with their capabilities. Then, in the next section the existing limitations of Translation-based embedding models emphasized in recent works are reviewed. These limitations will be reinvestigated in the light of \textit{score} and \textit{loss} functions in the section \ref{OurModel}.  
%Finally, we review the existing embedding models that incorporate relation patterns in their embeddings. 

\textbf{TransE}~\citep{bordes2013translating} is one of the earlier KGE models which is efficient in both time and space. The score function of TransE is defined as:
$f_r(h,t) = \| \textbf{h} + \textbf{r} - \textbf{t} \|.$

\textbf{TransH}~\citep{transH} projects each entity ($\textbf{e}$) to the relation space ($\textbf{e}_{\perp} = \textbf{e} - \textbf{w}_r \textbf{e} \textbf{w}_r^T$). The score function is defined as $f_r(h,t) = \| \textbf{h}_{\perp} + \textbf{r} - \textbf{t}_{\perp} \|$. 
TransH can encode reflexive, one-to-many, many-to-one and many-to-many relations. However, recent theories \citep{kazemi2018simple} prove that encoding reflexive results in encoding the both symmetric and transitive which is undesired. 

\textbf{TransR}~\citep{transr} projects each entity ($\textbf{e}$) to the relation space by using a matrix provided for each relation ($\textbf{e}_{\perp} = \textbf{e} \textbf{M}_r, \, \textbf{M}_r \in R^{d_e \times d_r}$). TransR uses the same scoring function as TransH. 

%TransR embeds entities and relations into two different spaces. The intuition is that entities have multiple aspects and relations focus on various aspects. Therefore, entities and relations should be embedded in two different spaces. TransR suffers the same problem mentioned for TransH. 

\textbf{TransD}~\citep{transD2015} provides two vectors for each individual entities and relations ($\textbf{h}, \textbf{h}_p, \textbf{r}, \textbf{r}_p, \textbf{t}, \textbf{t}_p$). Head and tail entities are projected by using the following matrices:

$
    \textbf{M}_{rh} = \textbf{r}_p^T \textbf{h}_p + \textbf{I}^{m \times n},  
     \textbf{M}_{rt} = \textbf{r}_p^T \textbf{t}_p + \textbf{I}^{m \times n}
$

The score function of TransD is similar to the score function of TransH.  
%TransD can better capture the diversity of both entities and relations while it has less number of parameters comparing to TransR \cite{transD2015}. 

\textbf{RotatE}~\citep{sun2019rotate} rotates the head to the tail entity by using relation. RotatE embeds entities and relations in Complex space. By inclusion of constraints on the norm of entity vectors, the model would be degenerated to TransE. The scoring function of RotatE is 
$
    f_r(h,t) = \| \textbf{h} \circ \textbf{r} - \textbf{t} \|, 
$
where $\textbf{h}, \textbf{r}, \textbf{t} \in C^d,$ and $\circ$ is element-wise product. RotatE obtains the state-of-the-art results using very big embedding dimension (1000) and a lot of negative samples (1000). 

\textbf{TorusE}~\citep{toruse2018} fixes the problem of regularization in TransE by applying translation on a compact Lie group. The model has several variants including mapping from torus to Complex space. In this case, the model is regarded as a very special case of RotatE \cite{sun2019rotate} that applies rotation instead of translation in the target Complex space. According to \cite{sun2019rotate}, TorusE is not defined on the entire Complex space. Therefore, it has less representation capacity. TorusE needs a very big embedding dimension (10000 as reported in \cite{toruse2018}) which is a limitation.

%\subsection{Relation Pattern Encoded Knowledge Graph Embedding Models}
%Relation patterns such as reflexive, symmetric, transitive, etc. are some of background knowledge that can be used on the top of triples during learning. They further improve performance of an embedding model. 
%Many existing embedding models encode the relational patterns either implicitly or explicitly \cite{sun2019rotate}. \newcite{minervini2017regularizing} encode inverse and equivalence relations in TransE, DistMult and ComplEx by adding regularization terms to the objective function. \newcite{manabe2018data} include L1 regularizer in the  ComplEx objective to promote symmetry or antisymmetry of the score function. RotatE \cite{sun2019rotate} provides a very spacial setting for embedding (i.e., a very big embedding dimension with a lot of negative samples) to enable the model to learn symmetric, antisymmetric, inverse and composition implicitly from data. SimplE \cite{kazemi2018simple} captures symmetric, antisymmetric and inverse patterns by weight tying in the model. 
%Other related works that encode relational patterns can be found in \cite{ruge,SimpleConst,lifted2016,minervini2017adversarial,logicconsist2018knowledge,rule2017training}.
\section{The Main Limitations Of Translation-based Embedding models}\label{lim}
Here we review the six limitations of translation-based embedding models in encoding relation patterns (e.g., reflexive, symmetric) mentioned in the literature: \cite{transH,kazemi2018simple,expressive, sun2019rotate}. 
% We later in the paper prove that these limitations are in fact mitigated by a proper \textit{loss} function and a proper modification on \textit{score} function. 

\textbf{Limitation L1}: \cite{transH}: TransE cannot encode reflexive relations when relation vector is non-zero. 

\textbf{Limitation L2} \cite{expressive}: if TransE encodes a relation $r$, which is neither reflexive nor irreflexive the following equations should be held simultaneously: 
$
\textbf{h}_1 + \textbf{r} = \textbf{h}_1 , 
\textbf{h}_2 + \textbf{r} \neq  \textbf{h}_2.
$
Therefore, both $\textbf{r} = \textbf{0}, \textbf{r} \neq \textbf{0}$ should be held, which result in contradiction. In this regard, TransE cannot encode a relation which is neither reflexive nor irreflexive.

\textbf{Limitation L3} \cite{sun2019rotate}: If relation $r$ is symmetric, the following equations should be held:
$\textbf{h} + \textbf{r} = \textbf{t}$ and  $\textbf{t} + \textbf{r} =  \textbf{h}$.
Therefore, $\textbf{r} = \textbf{0}$ and so all entities appeared in head or tail parts of training triples will have the same embedding vectors which is undesired. Therefore, TransE cannot properly encode symmetric relation when $\textbf{r} \neq \textbf{0}$. 

The following limitations are held for TransE, FTransE \cite{ftanse2016}, STransE \cite{stranse2016}, TransH and TransR.

\textbf{Limitation L4} \cite{kazemi2018simple}: if a relation $r$ is reflexive on $\Delta \in \mathcal{E},$ where $\mathcal{E}$ is the set of all entities in the KG, $r$ must also be symmetric. % on $\Delta.$ 

\textbf{Limitation L5} \cite{kazemi2018simple}: if $r$ is reflexive on $\Delta \in \mathcal{E},$ $r$ must also be transitive. % on $\Delta.$

\textbf{Limitation L6} \cite{kazemi2018simple}: if entity $e_1$ has relation $r$ with every entity in $\Delta \in \mathcal{E}$ and entity $e_2$ has relation $r$  with one of entities in $\Delta$, then $e_2$ must have the relation $r$ with every entity in $\Delta$.

\section{Our Model}\label{OurModel}

TransE and its variants underperform compared to other
embedding models due to their limitations we iterated in Section \ref{lim}.
In this section, we reinvestigate the limitations. 
We show that the corresponding theoretical proofs are inaccurate because the effect of loss function is ignored. So we propose new theories and prove that each of the limitations of TransE are resolved by revising either the \textit{scoring} function or 
the \textit{loss}. In this regard, we consider several loss functions and their effects on the boundary of the TransE scoring function. For each of the loss functions, we pose theories corresponding to the limitations. 
we additionally 
investigate the limitations of TransE using each of the loss functions while translation is performed in Complex space (TransComplEx).
%prove that
%performing translation in complex space (TransComplEx) further mitigates the limitations while the efficiency in memory and time are preserved.  
%propose a new variant of TransE, i.e.\ TransComplEx. 
TransComplEx with a proper selection of loss function further mitigates the limitations as we discuss as follows.
% and significantly outperforms translation-based embedding models and gets state-of-the-art performance. 

\subsection{TransComplEx: Translational Embedding Model in Complex Space}
%TransE and many other embedding models provide an embedding vector for each entity in a KG. In other words, regardless of being object or subject, each entity has one vector representation. In contrast, other embedding models such a CP decomposition \cite{cpdecomp} provide two vector representations for each entity to be used when the entity is subject or object. The former ignore the role of an entity as subject or object which is a limitation. The later learn two vectors independently which results in information loss.

%To clarify the negative effect of using two vectors for each entity, the following example is provided. Given the example $(A, Like, Juventus)$, $(Juventus, hasPlayer, C. Ronaldo),$ that $ C. Ronaldo$ plays for $Juventus$ may affect the person $A$ to like the team. This type of information is difficult to be captured by models that provide two independent vector representations for each entity (e.g., in the example, $Juventus$ is appeared as subject and object with two different embedding vectors) \cite{kazemi2018simple}. This example motivates to use Complex vector for each entity as an embedding vector. When an entity is appeared as subject, its original vector is used ($\textbf{h} = \textbf{e}$). In the case of using the entity as object, its conjugate is used ($\textbf{t} = \bar{\textbf{e}}$.) Therefore, the model provides same vectors for each entity regardless of being subject or object while it makes difference between the role of an entity as subject or object by using conjugate of the entity vector.
Inspired by \cite{trouillon2016complex}, in this section we propose TransComplEx that translates head entity vector to the conjugate of tail entity vector using relation vector in Complex space. 
%In this regard and inspired by \cite{trouillon2016complex}, we propose a Complex version of TransE, i.e. TransComplEx, that applies translation in Complex space. 
%In the next part, we prove that TransComplEx has less limitations in encoding relation patterns comparing to the TransE. 
The score function is defined as follows:

\begin{equation}
    f_r(h,t) = \| \textbf{h} + \textbf{r} - \bar{\textbf{t}}\|
\end{equation}

where $\textbf{h}, \textbf{r}, \textbf{t} \in \mathcal{C}^d$ are complex vectors i.e., each elements of the vectors is a complex number. For example, the $i$-th element of the vector $\textbf{h}$ is denoted by $h_i = Re(h_i) + Im(h_i)$. 
Respectively, $Re(.), Im(.)$ denote real and imaginary parts of a complex number. The complex vector $\textbf{h}$ contains real and imaginary vectors parts i.e.\ $\textbf{h} = Re(\textbf{h}) + Im(\textbf{h})$. $\bar{\textbf{t}} = Re(\textbf{t}) - Im(\textbf{t})$ is conjugate of the complex vector $\textbf{t}.$

\textbf{Advantages of TransComplEx:} 

i) Comparing to TransE and its variants, TransComplEx has less limitations in encoding different relation patterns. The theories and proofs are provided in the next part.

ii) Using conjugate of tail vector in the formulation enables the model to make difference between the role of an entity as subject or object. This cannot be properly captured by TransE and its variants.

iii) Given the example $(A, Like, Juventus)$, $(Juventus, hasPlayer, C. Ronaldo),$ that $ C. Ronaldo$ plays for $Juventus$ may affect the person $A$ to like the team. This type of information cannot be properly captured by models such as CP decomposition \cite{cpdecomp} where two independent vectors are provided \cite{kazemi2018simple} for $Juventus$ (for subject and object). In contrast, our model uses same real and imaginary vectors for $Juventus$ when it is used as subject or object. Therefore, TransComplEx can properly capture dependency between the two triples with the same entity used as subject and object. 

iiii) ComplEx \cite{trouillon2016complex} has much more computational complexity comparing to TransComplEx because it needs to compute eight vector multiplications to obtain score of a triple while our model only needs to do four vector summation/subtractions. 
In the experiment section, we show that TransComplEx outperforms ComplEx on various dataset.

%\textbf{Remark:} ComplEx computes the score of a triple as follows \cite{trouillon2016complex}:

%$f_r(h,t) = \, <Re(\textbf{r}),Re(\textbf{h}),Re(\textbf{t})> + \\ %<Re(\textbf{r}),Im(\textbf{h}),Im(\textbf{t})> + \\ %<Im(\textbf{r}),Re(\textbf{h}),Im(\textbf{t})> - \\ %<Im(\textbf{r}),Im(\textbf{h}),Re(\textbf{t})>.$

%Regarding the equation, several multiplications are done to compute the score of a triple for ComplEx whereas two summations and subtractions are done to compute the score of a triples for TransComplEx. Therefore, TransComplEx has less computational complexity comparing with ComplEx

% In the next part, we prove that TransComplEx provides a more flexible model with less limitations comparing to the TransE. 

\subsection{Reinvestigation of the Limitations of Translation-based Embedding Models}
%Although KGEs use ranking instead of classification, researchers \cite{kazemi2018simple,sun2019rotate} who work on theories of KGE consider a fix number for score of positive triples (e.g., $0$) to prove their theories.

The aim of this part is to analyze the limitations of Translation-based embedding models (including TransE and TransComplEx) by considering the effect of both \textit{score} and \textit{loss} functions. 
Different loss functions provide different upper-bound and lower-bound for positive and negative triples scores, respectively. Therefore, the loss functions affect the limitations of the models to encode relation patterns. To investigate the limitations, we redefine the conditions that a triple is considered as positive or negative by defining upper-bound and lower-bound for the scores. 

Lets $f_r(h,t)$, $f_r(h^{'},t^{'})$ be the scores of a positive ($h,r,t$) and negative ($h^{'},r,t^{'}$) triples respectively. The negative triple ($h^{'},r,t^{'}$) is generated by corruption of either head or tail of the triple ($h,r,t$) as mentioned in \cite{bordes2013translating}.
Four conditions are defined as follows:
\begin{equation}
\begin{cases}
%Condition & Positive & Negative & $\gamma_1, \gamma_2$\\
(a)  \, f_r(h,t) = \gamma_1,  \,    f_r(h^{'},t^{'}) \geq \gamma_2  ,  \, \gamma_1 = 0, \, \gamma_2 > 0  \\
(b) \, f_r(h,t) = \gamma_1 \, f_r(h^{'},t^{'}) \geq \gamma_2 \, \gamma_2 > \gamma_1 > 0  \\ 
(c)   \, f_r(h,t) \leq \gamma_1  \, f_r(h^{'},t^{'}) \geq \gamma_2 \, \gamma_2 > \gamma_1 > 0  \\ 
(d) \, f_r(h,t) \leq \gamma_{1(h,r,t)}, \, f_r(h^{'},t^{'}) \geq \gamma_{2(h,r,t)}, \,\\ \gamma_{2(h,r,t)} > \gamma_{1(h,r,t)} > 0   \\ 
\end{cases}
\end{equation}

%\begin{equation}
%\begin{cases}
%(a)f_r(h,t) = \gamma_1, \, f_r(h^{'},t^{'}) \geq \gamma_2 , \, \gamma_1 = 0,\\ \gamma_2 > 0\\
%(b)f_r(h,t) = \gamma_1, \, f_r(h^{'},t^{'}) \geq \gamma_2 , \, \, \gamma_2 > \gamma_1 > 0\\
%(c)f_r(h,t) \leq \gamma_1, \, f_r(h^{'},t^{'}) \geq \gamma_2 , \, \, \gamma_2 > \gamma_1 > 0\\
%(d)f_r(h,t) \leq \gamma_{1(h,r,t)}, \, f_r(h^{'},t^{'}) \geq \gamma_{2(h,r,t)},\\ \, \, \gamma_{2(h,r,t)} > \gamma_{1(h,r,t)} > 0.
%\end{cases}  
%end{equation}

\begin{figure}[!h]
\centering
\includegraphics[width=0.45\textwidth]{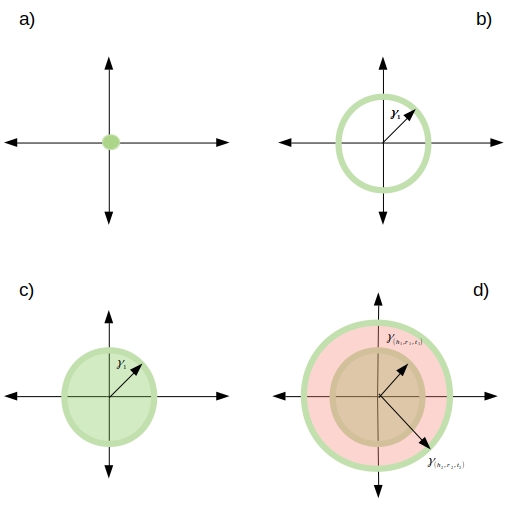} %need a better resolution: width=1.0\textwidth
\caption{The region of truth for a triple: A triple is positive if (a) its residual vector (i.e., $\epsilon = \textbf{h} + \textbf{r} - \textbf{t}$) becomes $\textbf{0}$ (b) its residual vector (i.e., $\epsilon$) lies on the border of a sphere with radius $\gamma_1,$ (c) its residual vector (i.e., $\epsilon$) lies inside of a sphere with radius $\gamma_1$, (d) its residual vector (i.e., $\epsilon_{(h_1,r_1,t_1)}$) lies inside of a sphere with radius $\gamma_{(h_1,r_1,t_1)}.$}.
\label{fig:F1}
\end{figure}

\begin{figure}[!h]
\centering
\includegraphics[width=0.45\textwidth]{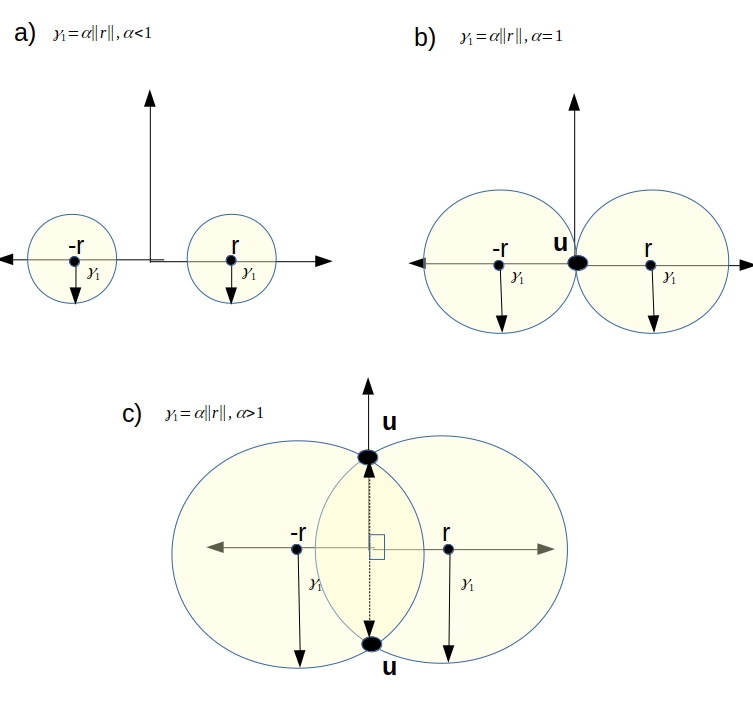} 
\caption{Necessity condition for encoding symmetric relation: (a) when $\alpha < 1, $ the model cannot encode symmetric relation.There is not any common points between two hyperspheres). (b) when $\alpha = 1, $ the intersection of two hyperspheres is a point. \textbf{u} = \textbf{0} means embedding vectors of all entities should be same. Therefore, symmetric relation cannot be encoded. (c) if $\alpha > 1, $ symmetric relation can be encoded because there are several points which are intersection of two hyperspheres.}
\label{fig:F2}
\end{figure}

Figure \ref{fig:F1} visualizes different conditions mentioned above. 
The condition (a) indicates a triple is positive if $\textbf{h} + \textbf{r} = \textbf{t}$ holds. It means that the length of \textit{residual vector} i.e., $\epsilon = \textbf{h} + \textbf{r} - \textbf{t},$ is zero. It is the most strict condition that expresses being positive. Authors in \cite{sun2019rotate,kazemi2018simple} consider this condition to prove their theories.  

Condition (b) considers a triple to be positive if its residual vector lies on a hyper-sphere with radius $\gamma_1.$ It is less restrictive than the condition (a) which considers a point to express being positive.  
The optimization problem that satisfies the conditions (a) ($\gamma_1 = 0$) and (b) ($\gamma_1 > 0$) is as follows:

\begin{align} 
\begin{cases}
\min_{\xi_{h,t}} \sum_{(h,r,t) \in S^+}  {\xi_{h,t}}^2 
\\ 
 \text{s.t.}\\
f_{r}(h,t) = \gamma_1 , \, (h,r,t) \in S^+ \\
f_{r}(h^{'}, t^{'}) \geq \gamma_2 - {\xi_{h,t}} , \, (h^{'}, r, t^{'}) \in S^-  \\
\xi_{h,t} \geq 0
\end{cases}
\label{mainOptab}
\end{align}

where $S^+, S^-$ are the set of positive and negative samples. The loss function that satisfies the conditions (a) ($\gamma_1 = 0$) and (b) ($\gamma_1 > 0$) is:
\begin{equation}
\begin{split}
\mathcal{L}_{a|b} = \sum_{(h,r,t) \in S^+} \lambda_1 \|f_{r}(h,t) - \gamma_1\| \, + \\ 
\lambda_2\, \max(\gamma_2 - f_{r}(h^{'},t^{'}),0).
\end{split}
\label{loss10}
\end{equation} 

Condition (c) considers a triple to be positive if its residual vector lies inside a hyper-sphere with radius $\gamma_1.$ The optimization problem that satisfies the condition (c) is as follows \citep{Nayyeri2019}: 

\begin{align} 
\begin{cases}
\min_{\xi_{h,t}} \sum_{(h,r,t) \in S^+}  {\xi_{h,t}}^2 
\\ 
 \text{s.t.}\\
f_{r}(h,t) \leq \gamma_1 , \, (h,r,t) \in S^+ \\
f_{r}(h^{'}, t^{'}) \geq \gamma_2 - {\xi_{h,t}} , \, (h^{'}, r, t^{'}) \in S^-  \\
\xi_{h,t} \geq 0
\end{cases}
\label{mainOptab}
\end{align}

The loss function that satisfies the condition (c) is as follows \cite{Nayyeri2019}:
\begin{equation}\label{loss11}
\begin{split}
\mathcal{L}_c =  \sum_{(h,r,t) \in S^+} \lambda_1 \max(f_{r}(h,t) - \gamma_1,0)\,  + \\ 
\lambda_2\, \max(\gamma_2 - f_{r}(h^{'},t^{'}),0) 
\end{split}
\end{equation} 

\textbf{Remark:} The loss function which is defined in \cite{zhou2017learning} is slightly different from the loss \ref{loss10}. The former slides the margin while the later fixes the margin by inclusion of a lower-bound for the score of negative triples. The both losses put an upper-bound for scores of positive triples. 

Condition (d) is similar to (c). But it provides different $\gamma_1, \gamma_2$ for each triples. Using the condition (d), there is not a unique region of truth for all positive triples, rather for each triple ($h,r,t$) and its corresponding negative samples ($h^{'},r,t^{'}$) there are triple specific region of truth and falsity. 
Margin ranking loss \citep{bordes2013translating} satisfies the condition (d). The loss is defined as:
\begin{align}\label{loss12}
    \mathcal{L}_d = \sum \sum \,
    [f_r(h,t) + \gamma - f_r(h^{'}, t^{'})]_+
\end{align}
where $[x]_+=\max(0,x)$.
Considering the conditions (a), (b), (c) and (d), we investigate the limitations L1 ,..., L6. We prove that existing theories are invalid under some conditions. During the following investigations of the limitations, we assume that the relation vectors shouldn't be null because the null vector for relation results same embedding vectors for entities appeared in head and tail parts when conditions (a) is used.

\textbf{Limitation L1:} 
\textit{Lemma 1:} Let assumption (a) holds, then TransE and TransComplEx cannot infer a reflexive relation pattern with non-zero relation vector. 
With assumptions (b), (c) and (d), however, this is not true anymore and the models can infer reflexive relation patterns  

\textit{Proof:} the proofs are provided in the supplementary material file. 

\newcommand{\comment}[1]{}

\comment{
1) Let $r$ be a reflexive relation and condition a) holds. For TransE, we have 
$
    \textbf{h} + \textbf{r} - \textbf{h} = \textbf{0}.
$

Therefore, the relation vector collapses to a null vector ($\textbf{r} = \textbf{0}).$

For TransComplEx with condition a), the following equations are obtained:
$
    Re(\textbf{r}) = \textbf{0}, 
     Im(\textbf{r}) = -2 \textbf{h}.
$
Therefore, all entities will have the same real and imaginary parts.

2) Let (b) holds. For a reflexive relation $r$, we have
$
    \| Re(\textbf{r}) + 2 Im(\textbf{h})\| = \gamma_1, 
$
which is not contradiction. 
For TransE, we get $\|\textbf{r}\|=\gamma_1.$
Therefore, TransE with condition (b) is capable of encoding reflexive relation. 
The condition (b) is a special case of (c) , (d). Therefore, with the same token, it is proved that TransE and TransComplEx can encode reflexive relations.
}

\textbf{Limitation L2:} 
\textit{Lemma 2:} 1) TransComplEx can infer a relation pattern which is neither reflexive nor irreflexive with condition (b), (c) and (d). 2) TransE cannot infer the relation pattern which is neither reflexive nor irreflexive. 

\comment{

\textit{proof:} 
1) Let the relation $r$ be neither reflexive nor irreflexive. There exists two triples $(e_1,r,e_1), (e_2,r,e_2)$ that are positive and negative respectively. Therefore the following inequalities hold:

\begin{align}
\begin{cases}
    \| \textbf{e}_1 + \textbf{r} - \Bar{\textbf{e}_1} \| \leq \lambda_1, \\
    \| \textbf{e}_2 + \textbf{r} - \Bar{\textbf{e}_2} \| \geq \lambda_2.
\end{cases}
\label{R2}
\end{align}

Equation \ref{R2} is rewritten as follows:

$
    \| Re(\textbf{r}) + Im(\textbf{r}) + 2 Im(\textbf{e}_1)\| \leq \gamma_1, \\
    \| Re(\textbf{r}) + Im(\textbf{r}) + 2 Im(\textbf{e}_2)\| \geq \gamma_2, \\
$
For TransE in real space,  $\| Re(\textbf{r})\| \leq \gamma_1$ and $\| Re(\textbf{r}) \| \geq \gamma_2$ cannot be held simultaneously. Therefore, TransE in real space cannot encode a relation  which is neither reflexive nor irreflexive. In contrast, TransE in complex space can encode the relation by proper assignment of imaginary parts of entities. Therefore, theoretically TransComplEx can infer a relation which is neither reflexive nor irreflexive. 
}

\textbf{Limitation L3:} 
\textit{Lemma 3}: 1) TransComplEx can infer symmetric patterns with condition (a), (b), (c) and (d). 2) TransE cannot infer symmetric patterns with condition (a) with non-zero vector for relation. 3) TransE can infer a relation pattern which is symmetric with conditions (b). 

\textit{Proof:} proof of 1), 2) and 3) are included in the supplementary material.

\comment{
1), 2) Let $r$ be a symmetric relation and (a) holds. We have
$
    \textbf{h} + \textbf{r} = \bar{\textbf{t}}, 
    \textbf{t} + \textbf{r} = \bar{\textbf{h}}.
$
Trivially, we have
$\\
    Re(\textbf{h}) + Re(\textbf{r}) = Re(\textbf{t}), \\
    Re(\textbf{t}) + Re(\textbf{r}) = Re(\textbf{h}), \\
    Im(\textbf{h}) + Im(\textbf{r}) = -Im(\textbf{t}), \\
    Im(\textbf{t}) + Im(\textbf{r}) = -Im(\textbf{h}), \\
$
For TransE in real space, there is
\begin{align*}
    Re(\textbf{h}) + Re(\textbf{r}) = Re(\textbf{t}), \\
    Re(\textbf{t}) + Re(\textbf{r}) = Re(\textbf{h}), 
\end{align*}
Therefore, $Re(\textbf{r}) = \textbf{0}.$ It means that TransE cannot infer symmetric relations with condition (a). For TransComplEx, additionally we have
\begin{align*}
    Im(\textbf{h}) + Im(\textbf{r}) = -Im(\textbf{t}), \\
    Im(\textbf{t}) + Im(\textbf{r}) = -Im(\textbf{h}), 
\end{align*}
It concludes $Im(\textbf{h}) + Im(\textbf{r}) + Im(\textbf{t}) = \textbf{0}$. Therefore, TransE in complex space with condition (a) can infer symmetric relation. 
Because (a) is an special case of (b) and (c), TransComplEx can infer symmetric relations in all conditions. 
}
3) For TransE with condition (b), there is 
\begin{align}
    \| \textbf{h} + \textbf{r} - \textbf{t} \| = \gamma_1, \\
    \| \textbf{t} + \textbf{r} - \textbf{h} \| = \gamma_1. 
    \label{910}
\end{align}
The necessity condition for encoding symmetric relation is
$\| \textbf{h} + \textbf{r} - \textbf{t} \| = \| \textbf{t} + \textbf{r} - \textbf{h} \|.$ This implies 
$
    \|h \| cos(\theta_{h,r}) = \|t \| cos(\theta_{t,r}).
$
Let $h-t = u$, by definition we have 
$
    \| \textbf{u} + \textbf{r}  \| = \gamma_1, 
    \| \textbf{u} - \textbf{r} \| = \gamma_1. 
    \label{15}
$

Let $\gamma_1 = \alpha \|r\|$. We have
\begin{align}
\begin{cases}
    \|\textbf{u}\|^2 + (1-\alpha^2) \|\textbf{r}\|^2 = - 2 \langle \textbf{u},\textbf{r}\rangle \\ 
    \|\textbf{u}\|^2 + (1-\alpha^2) \|\textbf{r}\|^2 =  2 \langle \textbf{u},\textbf{r}\rangle
\end{cases}
\label{168}
\end{align}
Regarding \ref{168}, there is

$\|\textbf{u}\|^2 + (1-\alpha^2) \|\textbf{r}\|^2 = -(|\textbf{u}\|^2 + (1-\alpha^2) \|\textbf{r}\|^2). $
$ \rightarrow \|\textbf{u}\|^2 = (\alpha^2 - 1) \|\textbf{r}\|^2.$

To avoid contradiction,  $\alpha > 1$. If $\alpha > 1$ we have $ \cos(\theta_{u,r}) = \pi /2 .$
Therefore, TransE can encode symmetric pattern with condition (b), if $\gamma_1 = \alpha \|r\|$ and  $\alpha > 1$. Figure \ref{fig:F2} shows different conditions for encoding symmetric relation. 

\comment{
\begin{figure}[!h]
\centering
\includegraphics[width=0.45\textwidth]{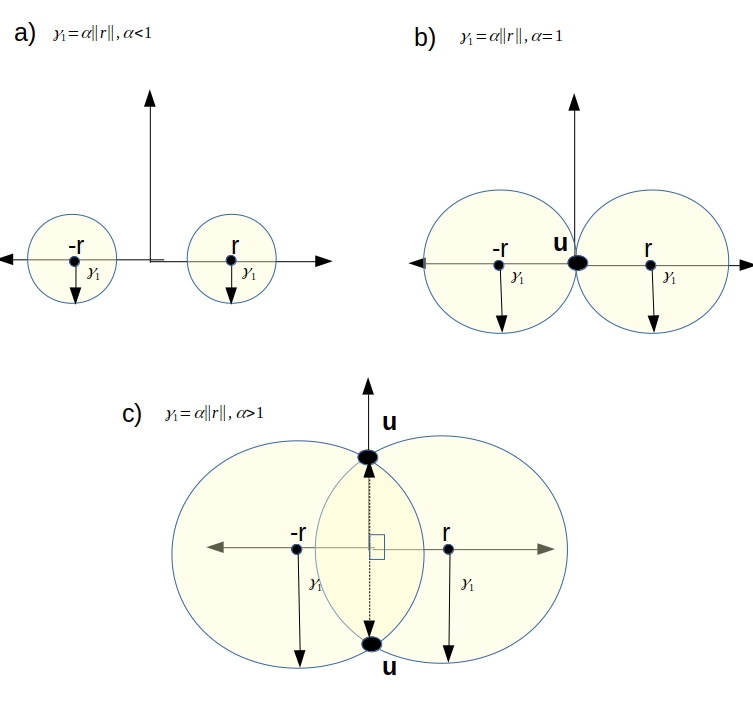}
\caption{Necessity condition for encoding symmetric relation: (a) when $\alpha < 1, $ the model cannot encode symmetric relation.There is not any common points between two hydrspheres). (b) when $\alpha = 1, $ the intersection of two hyperspheres is a point. \textbf{u} = \textbf{0} means embedding vectors of all entities should be same. Therefore, symmetric relation cannot be encoded. (c) if $\alpha > 1, $ symmetric relation can be encoded because there are several points which are intersection of two hyperspheres.}
	\label{fig:F2}
\end{figure}
}
\textbf{Limitation L4:}
\textit{Lemma 4:} 1) Let (a) holds. Limitation L4 holds for both TransE and TransComplEx. 2) Limitation L4 is not valid when assumptions (b), (c) and (d) hold. 

%\textit{Proof:} The proofs are included in the supplementary file.

\comment{
1) The proof of the lemma with condition (a) for TransE is mentioned in the paper \cite{kazemi2018simple}. For TransComplEx, the proof is trivial. 
2) Now, we prove that the limitation L4 is not valid when (b) holds. 

Let condition (b) holds and relation $r$ be reflexive, we have
$    \| \textbf{e}_1 + \textbf{r} -\textbf{e}_1\| = \gamma_1,
    \| \textbf{e}_2 + \textbf{r} -\textbf{e}_2\| =\gamma_1.
$

Let $\| \textbf{e}_1 + \textbf{r} - \textbf{e}_2 \| = \gamma_1$. To violate the limitation L4, the triple $(e_2, r, e_1)$ should be negative i.e.,

$\| \textbf{e}_2 + \textbf{r} - \textbf{e}_1 \| > \gamma_1,$

$\rightarrow \| \textbf{e}_2 + \textbf{r} - \textbf{e}_1 \|^2 > \gamma_1^2,$

$\rightarrow \| \textbf{e}_2 \|^2 + \| \textbf{e}_1 \|^2 + \| \textbf{r} \|^2 + 2 <\textbf{e}_2,\textbf{r}> - 2 <\textbf{e}_2,\textbf{e}_1> - 2 <\textbf{e}_1,\textbf{r}> \,\, > \gamma_1^2.$

Considering $\| \textbf{e}_1 + \textbf{r} - \textbf{e}_2 \| = \gamma_1$, we have

$
     <\textbf{e}_2,\textbf{r}> - <\textbf{e}_1,\textbf{r}> \,\, > 0, \\
     \rightarrow \, <\textbf{e}_2 - \textbf{e}_1,\textbf{r}> \,\, > 0, \\
     \rightarrow \, cos(\theta_{(\textbf{e}_2 - \textbf{e}_1), \textbf{r}}) \,\, > 0,
$

Therefore, the limitation L4 is not valid i.e., if a relation $r$ is reflexive, it may not be symmetric.
TransE is special case of TransComplEx and also condition (b) is special case of condition (c). Therefore using conditions (b), (c) and (d), the limitation L4 is not valid for TransE and TransComplEx.
}
\textbf{Limitation L5:} 
\textit{Lemma 5:}  1) Under condition (a), the limitation L5 holds for both TransE and TransComplEx. 2) Under conditions (b), (c) and (d), L5 is not valid for both TransE and TransComplEx. 

%\textit{proof}: The proofs are included in the supplementary material. 

\comment{
1) Under condition (a), equation $\textbf{h} + \textbf{r} - \textbf{t} = \textbf{0}$ holds. Therefore, according to the paper \cite{kazemi2018simple}, the model has the limitation L5. 

2) If a relation is reflexive, with condition (b), we have
$
    \| \textbf{e}_1 + \textbf{r} -\textbf{e}_1\| = \gamma_1,
    \| \textbf{e}_2 + \textbf{r} -\textbf{e}_2\| =\gamma_1.
$
Therefore, $\|r\|= \lambda_1.$
Let 
\begin{align}
    \begin{cases}
    \| \textbf{e}_1 + \textbf{r} -\textbf{e}_2\| = \gamma_1,\\
    \| \textbf{e}_2 + \textbf{r} -\textbf{e}_3\| =\gamma_1.
    \end{cases}
\label{22}
\end{align}

we need to show the following inequality wouldn't give contradiction:
$
    \| \textbf{e}_2 + \textbf{r} -\textbf{e}_3\|  >
\gamma_1.
$

From \ref{22} we have
$< \textbf{e}_2, (\textbf{e}_1 + \textbf{e}_2 + \textbf{e}_3) >  < 0,$ which is not contradiction. 

Therefore, with conditions (b) and (c), the limitation L5 is not valid for both TransE and TransComplEx. 
}

\textbf{Limitation L6:}
\textit{Lemma 6:} 1) With condition (a), the limitation L6 holds for both TransE and TransComplEx. 2) With conditions (b), (c) and (d), the limitation L6 doesn't hold for the models. 
\comment{
\textit{Proof}: 
1) With condition (a), the limitation L6 is proved in \cite{kazemi2018simple}.
2) Considering the assumption of L6 and the condition (b), we have

\begin{align}
    \begin{cases}
    \| \textbf{e}_1 + \textbf{r} -\textbf{s}_1\| = \gamma_1,\\
    \| \textbf{e}_1 + \textbf{r} -\textbf{s}_2\| =\gamma_1.\\
    \| \textbf{e}_2 + \textbf{r} -\textbf{s}_1\| =\gamma_1.
    \end{cases}
    \label{24}
\end{align}

We show the condition that $\| \textbf{e}_2 + \textbf{r} -\textbf{s}_2\| > \gamma_1$ holds.

Substituting \ref{24} in  $\| \textbf{e}_2 + \textbf{r} -\textbf{s}_2\| > \gamma_1$, we have 

$cos(\theta_{(s_1-s_2),(e_1-e_2)}) < 0.$ Therefore, the limitation L6 is not valid with condition (b), (c) and (d). 

\begin{figure}[!h]
\centering
\includegraphics[width=0.45\textwidth]{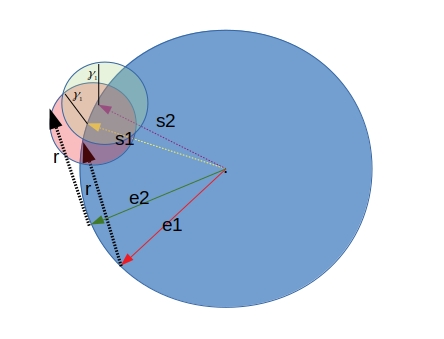} %need a better resolution: width=1.0\textwidth
\caption{Investigation of L6 with condition (c): The limitation is not valid, because the triple ($e_2,r,s_2$) can get an score to be considered as negative while triples ($(e_1,r,s_1), (e_1,r,s_2), (e_2,r,s_1)$) are positive.}
	\label{fig:F1}
\end{figure}
}
\subsection{Encoding Relation Patterns in TransComplEx}
Most of KGE models learn from triples. Recent work incorporates relation patterns such as transitive, symmetric on the top of triples to further improve performance of models. For example, ComplEx-NNE+AER \cite{ding2018improving} encodes implication pattern in the ComplEx model. RUGE \cite{ruge} injects First Order Horn Clause rules in an embedding model. SimplE \cite{kazemi2018simple} captures symmetric, antisymmetric and inverse patterns by weight tying in the model. Inspired by \cite{minervini2017regularizing} and considering the score function of TransComplEx, in this part, we derive formulae for equivalence, symmetric, inverse and implication to be used as regularization terms in the optimization problem. Therefore, the model incorporates different relation patterns to optimize the embeddings.

\textbf{Symmetric:} In order to encode symmetric relation $r$, the following should be held: 
$$f_r(h,t) \Longleftrightarrow f_r(t,h),$$
Therefore the following algebraic formulae is proposed to encode the relation:
$\| f_r(h,t) - f_r(t,h) \| = 0.$ According to the definition of score function of TransComplEx, we have the following algebraic formulae:
$\mathcal{R_S} = \| Re(\textbf{h}) - Re(\textbf{t}) \| = 0.$
Using similar argument for symmetric, the following formulae are derived for transitive, composition, inverse and implication:

\textbf{Equivalence:} Let $p,q$ be equivalence relations i.e.,
$f_p(h,t) \Longleftrightarrow f_q(h,t).$
we obtain
$\mathcal{R_E} = \| \textbf{p} - \textbf{q} \| = 0.$

\textbf{Implication:} Let $p \rightarrow q,$
we obtain
$\mathcal{R_I} = \max( f_p(h,t) - f_q(h,t) , 0 ) = 0.$

\textbf{Inverse:} Let $r \longleftrightarrow r^{-1},$
we obtain
$\mathcal{R}_{In} = \| \textbf{r} - \textbf{r}^{-1} \|.$

Finally, the following optimization problem should be solved:
\begin{equation}\label{regularize}
    \min_{\theta} \, \, \mathcal{L} + \sum \eta_i \mathcal{R}_i
\end{equation}

where $\theta$ is embedding parameters, $\mathcal{L}$ is one of the losses \ref{loss10}, \ref{loss11} or \ref{loss12} and $\mathcal{R}$ is one of the derived formulae mentioned above.

\section{Experiments and Evaluations}\label{exper}

In this section, we evaluate performance of our model, TransComplEx, with different loss functions on link prediction task. The aim of the task is to complete the triple $(h,r,?)$ ($(?,r,t)$) by prediction of the missed entity $h$ or $t$. Filtered Mean Rank (MR), Mean Reciprocal Rank (MRR) and Hit@10 are used for evaluations \citep{wang2017knowledge, transr}.

\paragraph{Dataset.} We use two dataset extracted from Freebase \citep{bollacker2008freebase} (i.e., FB15K \citep{bordes2013translating} and FB15K-237 \citep{toutanova2015observed}) and two others extracted from WordNet~\citep{miller1995wordnet} (i.e.\ WN18 \citep{bordes2013translating} and WN18RR \citep{dettmers2018convolutional}). FB15K and WN18 are earlier dataset which have been extensively used to compare performance of KGEs. FB15K-237 and WN18RR are two dataset which are supposed to be more challenging after removing inverse patterns from FB15K and WN18. \cite{ruge} and \cite{ding2018improving} extracted different relation patterns from FB15K and WN18 respectively. The relation patterns are provided by their confidence level, e.g.\ $(a, BornIn, b) \xrightarrow{0.9} (a, Nationality, b)$. We drop the relation patterns with confidence level less than 0.8. Generally, we use 454 and 14 relation patterns for FB15K and WN18 respectively. We do grounding for symmetric and transitive relation patterns. Thanks to the formulation of score function, grounding is not needed for inverse, implication and equivalence. 

\paragraph{Experimental Setup.} We implement TransComplEx with the losses \ref{loss10}, \ref{loss11} and \ref{loss12} and TransE with the loss \ref{loss11} in Pytorch. Adagrad is used as an optimizer. We generate 100 mini-batches in each iteration. The hyperparameter corresponding to the score function is embedding dimension $d$. We add slack variables to the losses \ref{loss10} and \ref{loss11} to have soft margin as in \citep{Nayyeri2019}. The loss \ref{loss11} is rewritten as follows \cite{Nayyeri2019}:
\begin{equation}
    \begin{aligned}
    \min_{\xi_{h,t}^r} \sum_{(h,r,t) \in S^+} \lambda_0 {\xi_{h,t}^r}^2 + 
\lambda_1 \max(f_{r}(h,t) -
\gamma_1,0) + \\
\lambda_2\, \max(\gamma_2 - f_{r}(h^{'},t^{'}) - {\xi_{h,t}^r},0) 
    \end{aligned}
\end{equation}

We set $\lambda_1$ and $\lambda_2$ to one and search for the hyperparameters $\gamma_1 (\gamma_2 > \gamma_1)$ and $\lambda_0$ in the sets $\{0.1, 0.2, \ldots, 2\}$ and $\{0.01, 0.1, 1, 10, 100\}$ respectively. Moreover, we generate $\alpha \in \{1, 2, 5 , 10\}$ negative samples per each positive. The embedding dimension and learning rate are tuned from the sets $\{100, 200\}, \{0.0001 , 0.0005 ,0.001 , 0.005, 0.01\}$ respectively. All hyperparameters are adjusted by early stopping on validation set according to MRR. 
RPTransComplEx$\#$ denotes the TransComplEx model which is trained by the loss function $\#$ (\ref{loss10}, \ref{loss11}, \ref{loss12}). RP indicates that relation patterns are injected during learning by regularizing the derived formulae (see \ref{regularize}). TransComplEx$\#$ refers to our model trained with the loss $\#$ without regularizing relation patterns formulae. The same notation is used for TransE$\#$. The optimal configurations for 
RPTransComplEx\ref{loss10} are $d =200 , \lambda_0 =100 , \gamma_1 =0.4 , \gamma_2 =0.5 , \alpha =10 $ for FB15K,  $d =200 , \lambda_0 =100 , \gamma_1 =1.5 , \gamma_2 =2 , \alpha =10 $ for FB15K-237,  $d =200 , \lambda_0 =100 , \gamma_1 =1 , \gamma_2 =2 , \alpha =10 $ for WN18; 
for RPTransComplEx\ref{loss11} are $d =200 , \lambda_0 =10 , \gamma_1 =0.4 , \gamma_2 =0.5 , \alpha =10 $ for FB15K,  $d = 200, \lambda_0 =100 , \gamma_1 =1.5 , \gamma_2 =2 , \alpha = 10$ for FB15K-237,  $d = 200, \lambda_0 = 100, \gamma_1 = 0.6, \gamma_2 = 1.7, \alpha = 2$ for WN18; 
for RPTransComplEx\ref{loss12} are $d = 200, \gamma = 5, \alpha =10$ for FB15K,  $d =200, \gamma =10, \alpha =10$ for FB15K-237,  $d =200, \gamma =10, \alpha =10$ for WN18; 
for TransComplEx\ref{loss11} are $d =200 , \lambda_0 =10 , \gamma_1 =0.4 , \gamma_2 =0.5 , \alpha =10 $ for FB15K,  $d = 200, \lambda_0 =100 , \gamma_1 =1.5 , \gamma_2 =2 , \alpha = 10$ for FB15K-237,  $d = 200, \lambda_0 = 100, \gamma_1 = 0.6, \gamma_2 = 1.7, \alpha = 2$ for WN18,  $d = 200, \lambda_0 = 1, \gamma_1 = 1.6, \gamma_2 = 2.7, \alpha = 2$ for WN18RR, 
for TransE\ref{loss11} are $d =200 , \lambda_0 =10 , \gamma_1 =0.4 , \gamma_2 =0.5 , \alpha =10 $ for FB15K,  $d =200 , \lambda_0 =100 , \gamma_1 =0.4 , \gamma_2 =0.5 , \alpha =10 $ for FB15K-237,  $d =200 , \lambda_0 =1 , \gamma_1 =1 , \gamma_2 =2 , \alpha =10 $ for WN18,  $d = 200, \lambda_0 = 1, \gamma_1 = 0.6 , \gamma_2 = 1.7, \alpha = 2$ for WN18RR.

\begin{table*}[!h]
    \footnotesize
    \centering
    \renewcommand\tabcolsep{2.0pt}
    \resizebox{0.8 \textwidth}{!}{
      \begin{tabular}{lcccccccccc}  
        \toprule  
        %\multirow{}{}{}&  
        %\multicolumn{6}{c}{ FB15k}&\multicolumn{6}{c}{ WN18}\cr  
       % \cmidrule(lr){2-7} \cmidrule(lr){8-13}  
        &\multicolumn{4}{c}{FB15k}&\multicolumn{3}{c}{WN18}\cr 
            \cmidrule(lr){2-5} \cmidrule(lr){6-8}
            \multicolumn{3}{c}{}&\multicolumn{1}{c}{Hits}&\multicolumn{3}{c}{}&\multicolumn{1}{c}{Hits}\cr
            &MR&MRR&@10&&MR&MRR&@10\cr
        \midrule
TransE~\citep{bordes2013translating}& 125& -&47.1 & & 251& - & 89.2 \cr
%TransH (Unif)~\cite{transH}& 84& - &58.5 & & 303& - & 86.7 \cr
TransH (bern)~\citep{transH}*& 87& - &64.4 & &388 & - & 82.3 \cr
%TransR (Unif)~\cite{transr}& 78& - &65.5 & & 219& - & 91.7 \cr
TransR (bern)~\citep{transr}*& 77& - &68.7 & &225 & - & 92.0 \cr
%TransD (Unif)~\cite{transD2015}& 67& - &74.2 & & 229& - & 92.5 \cr
TransD (bern)~\citep{transD2015}*& 91& - &77.3 & & \textbf{212} & - &92.2 \cr

%TransE-RS (Unif)~\cite{transD2015}& 62& - &72.3 & & 348& - & 93.7 \cr
TransE-RS (bern)~\citep{cikmzhou2017learning}* & 63& - &72.1 & &371 & - &93.7 \cr

%TransH-RS (Unif)~\cite{cikmzhou2017learning}& 64& - &72.6 & & 389& - & 94.7 \cr
TransH-RS (bern)~\citep{cikmzhou2017learning}*& 77& - &75.0 & &357 & - &94.5 \cr

TorusE~\citep{torusJournal2019generalized}&- & 73.3 & 83.2 & & -& 94.7 & 95.4 \cr
TorusE(with WNP)~\citep{torusJournal2019generalized}&- & \textbf{75.1} & 83.5 & & -& \textbf{94.7} & 95.4 \cr
\midrule
R-GCN~\citep{rgcn}+ &-& 65.1 & 82.5 & &- & 81.4 & \textbf{95.5} \cr
ConvE~\citep{dettmers2018convolutional}++&51& 68.9 &85.1 & &504 & 94.2 &\textbf{95.5} \cr
ComplEx~\citep{trouillon2016complex}++& 106 & 67.5 & 82.6 & &543 & 94.1 &94.7 \cr
ANALOGY~\citep{liu2017analogical}++& 121 & 72.2 & 84.3 & &- & 94.2 & 94.7 
\cr
RotatE~\citep{sun2019rotate}& 48 & 69.0 & 86.1 & & 433 & \textbf{94.8} & \textbf{95.5} \cr

\midrule
SimplE~\citep{kazemi2018simple}&-& 72.7 & 83.8 & & - & 94.2 & 94.7 \cr
SimplE+~\citep{simpleplus}&-& 72.5 & 84.1 & &- & 93.7 & 93.9 \cr
PTransE~\citep{ptranse}& 58 & - & 84.6 & &- & - & - \cr
KALE~\citep{kalejoint}& 73 & 52.3 & 76.2 & & 241 & 53.2 & 94.4 \cr
RUGE~\citep{ruge}& 97 & 76.8 & 86.5 & & - & - & - \cr
ComplEx-NNE+AER~\citep{ding2018improving} & 116 & 80.3 & 87.4 & & 450 & 94.3 & 94.8 \cr

\midrule
RPTransComplEx\ref{loss10} & \textbf{38}  & 70.5 & 88.3 & & 451  & 92.7   & 94.8 & & & \cr
RPTransComplEx\ref{loss11} & \textbf{38} & 72.4 & \textbf{88.8} & & 275 & 92.4 & 95.4\cr
RPTransComplEx\ref{loss12} & 59& 61.7 &  82.2& & 547& 94.0 & 94.7 \cr 
TransComplEx\ref{loss11} & \textbf{38} & 68.2 & 87.5 & & 284 & 92.2 & \textbf{95.5} &&&\cr
TransE\ref{loss11} & 46 & 64.8 & 87.2 & &703  & 68.7& 94.5&&&\cr
        \bottomrule
    \end{tabular}       
    } % resizebox
    \caption{
    Link prediction results. 
    Rows 1-8: Translation-based models with no injected relation patterns.
    Rows 9-13: basic models with no injected relation patterns.
    Rows 14-18: models which encode relation patterns. 
    Results labeled with *, + and ++ are taken from  \protect\cite{cikmzhou2017learning},  \protect\cite{torusJournal2019generalized} and \protect\cite{akrami2018re}
    while the rest are taken from original papers/code.
    Dashes: results could not be obtained. }
    \label{tbl:AllResults}
\end{table*}

\begin{table*}[!h]
    \footnotesize
    
    \centering
    \renewcommand\tabcolsep{2.0pt}
    \resizebox{0.8 \textwidth}{!}{
      \begin{tabular}{lcccccccccc}  
        \toprule  
        
        %\multirow{}{}{}&  
        %\multicolumn{6}{c}{ FB15k}&\multicolumn{6}{c}{ WN18}\cr  
       % \cmidrule(lr){2-7} \cmidrule(lr){8-13}  
        &\multicolumn{4}{c}{FB15k-237}&\multicolumn{3}{c}{WN18RR}\cr 
            \cmidrule(lr){2-5} \cmidrule(lr){6-8}
            \multicolumn{3}{c}{}&\multicolumn{1}{c}{Hits}&\multicolumn{3}{c}{}&\multicolumn{1}{c}{Hits}\cr
            &MR&MRR&@10&&MR&MRR&@10\cr
        \midrule
TransE~\citep{bordes2013translating}+&- & 25.7& 42.0 & & - & 18.2 & 44.4 \cr
DistMult~\citep{bordes2013translating}+&- & 24.1& 41.9 & & - & 43.0 & 49.0 \cr
ComplEx~\citep{trouillon2016complex}+&- & 24.0& 41.9 & & - & 44.0 & 51.0 \cr
R-GCN~\citep{rgcn}+ &- & 24.8& 41.7 & & - & - & - \cr
ConvE~\citep{dettmers2018convolutional}+&- & 31.6& 49.1 & & - & 46.0 & 48.0 \cr
TorusE~\citep{torusJournal2019generalized}&- & 30.5& 48.4 & & - & 45.2 & 51.2 \cr
TorusE (with WNP)~\citep{torusJournal2019generalized}&- & 30.7& 48.5 & & - & 46.0 & 53.4 \cr
RotatE~\citep{sun2019rotate}&211 & 31.1 & 49.4 & & 4789 & \textbf{47.3} & \textbf{54.9} \cr
\midrule
RPTransComplEx\ref{loss10} &210 &27.7 & 46.4 & &-&-&-\cr
RPTransComplEx\ref{loss11} &226 & \textbf{31.9} & \textbf{49.5} & &-&-&-\cr
RPTransComplEx\ref{loss12} &216&25.3& 43.8& &-&-&-\cr
TransComplEx\ref{loss11} &223 &31.7 & 49.3&&4081&38.9&49.8\cr
TransE\ref{loss11} & \textbf{205} & 27.2 & 45.3 & & 3850  &20.0&47.5\cr
        \bottomrule
    
    \end{tabular}       
    } % resizebox
    \caption{
Link prediction results. 
    Rows 1-8: basic models with no injected relation patterns.
    Results labeled with + are taken from    \protect\cite{torusJournal2019generalized}
    while the rest are taken from original papers/code.
    Dashes: results could not be obtained. 
    }
    \label{tbl:AllResultsnewds}
\end{table*}

\paragraph{Results.} Table \ref{tbl:AllResults} presents comparison of TransComplEx and its relation pattern encoded variants (RPTransComplEx) with three classes of embedding models including Translation-based models (e.g.\ TransX, TorusE), relation pattern encoded models (e.g.\ RUGE, ComplEx-NNE+AER, SimplE, SimplE+), and other state-of-the-art embedding models (e.g.\ ConvE, ComplEx, ANALOGY). To investigate our theoretical proofs corresponding to the effect of \textit{loss} function, we train TransComplEx with different loss functions. As previously discussed, FB15K-237 and WN18RR are two more challenging dataset provided recently. Therefore, in order to have a better evaluation, Table \ref{tbl:AllResultsnewds} presents comparison of our models with state-of-the-art embedding methods on these two dataset. For WN18RR, we do not use any relation patterns to be encoded. The results labeled with "*", "+" and "++" are taken from \cite{cikmzhou2017learning}, \cite{torusJournal2019generalized} and \cite{akrami2018re} respectively. To have a fair comparison, we ran the code of RotatE \citep{sun2019rotate} in our setting e.g., embedding dimension 200 and 10 negative samples while the original paper reported the results of RotatE using a very big embedding dimension and a lot of negative samples (embedding dimension 1000 and 1000 negative samples).

\paragraph{Boosting techniques:}
There are several ways to improve the performance of embedding models: 1) designing a more sophisticated scoring
function, 2) proper selection of loss function, 3) using more negative samples 4) using negative sampling techniques, 5) enriching dataset (e.g., adding reverse triples). 
Among the mentioned techniques, we focus on the first and second ones and avoid using other techniques. We keep the setting used in \cite{trouillon2016complex} to have a fair comparison. 
Using other techniques can further improve the performance of every models including ours. For example, TransComplEx with embedding dimension 200 and 50 negative samples gets 52.2 for Hits@10. 

\paragraph{Dissuasion of Results.} According to the Table \ref{tbl:AllResults}, FB15K dataset part, PRTransComplEx trained by the loss \ref{loss11} significantly outperforms all Translation-based embedding models including the recent work TorusE. Note that TorusE is trained by embedding dimension 10000 while our model uses embedding dimension at most 200. Comparing to relation pattern encoded embedding models including recent works ComplEx-NNE+AER, RUGE, SimplE and SimplE+, our model outperforms them in the terms of MR and Hit@10. Moreover, the model significantly outperforms popular embedding models including ConvE and ComplEx. Regarding our theories, the loss \ref{loss11} has less limitations comparing to the loss \ref{loss10}. This is consistent with our theories where RPTransComplEx\ref{loss11} outperforms RPTransComplEx\ref{loss10}. TransComplEx without encoding relation patterns still obtains accuracy as good as state-of-the-art models. TransComplEx outperforms TransE while both are trained by the loss \ref{loss11} in the terms of MR, MRR and Hit@10 which is consistent with our theories (TransComplEx score function has less limitations than TransE). Regarding the results on WN18, the accuracy of TransComplEx is very close to the state-of-the-art models. Encoding relation patterns cannot improve the performance on WN18 because the models learn relation patterns from data well. The loss \ref{loss12} provides different upper-bounds and lower-bounds for the score of positive and negative triples respectively and also the margin can slide. Therefore, the accuracy would be degraded \cite{cikmzhou2017learning}. Generally, the loss \ref{loss11} gets better performance which is consistent to our theoretical results. As shown in the Table \ref{tbl:AllResultsnewds}, FB15K-237 part, with and without encoding relation patterns, TransComplEx trained by the loss \ref{loss11} outperforms all the baselines in terms of MRR and Hit@10. TransComplEx\ref{loss11} outperforms TransE\ref{loss11} showing the effectiveness of our proposed score function. Regarding WN18RR, TorusE has better performance comparing to our model. However, the results are obtained with a very big embedding dimension ($d = 10000$).

\comment{
\begin{tabular}{ |p{3.1cm}||p{0.5cm}|p{0.5cm}|p{0.5cm}| p{0.5cm}| p{0.5cm}|}
 \hline
 \multicolumn{6}{|c|}{Optimal Configuration} \\
 \hline
 Model & $d$ & $\gamma_1$ & $\gamma_2$ & $\alpha$ & $\lambda_0$\\
 \hline
 RPTransComplEx10   & AF    & AF &   04 & &\\
 RPTransComplEx11 &   AX  & AL   &24 & &\\
 RPTransComplEx12 &AL & AL &  00 & &\\
 TransComplEx11  &DZ & DZ &  01 & &\\
 TransE11 &   AS  & AS & 01 & &\\
 \hline
\end{tabular}
}

%Recent studies draw attention to main limitations of TransXs, encouraging KGR community to address these issues. In this study, we first shed light on existing theoretical proofs of the TransXs limitations in encoding relational patterns. We then show that these theories are inaccurate because the selection of loss function significantly affects those theories. In this regard, new theories are proposed to make a connection between TransX limitations and different loss functions. Second, we propose \name, a new variant of TransE that applies translation in complex space. TransComplEx is trained with a framework which has a fewer theoretical limitations in encoding relational patterns according to our theories. Experimental results show TransComplEx outperforms the state-of-the-art embedding models on standard benchmark datasets. 

\section{Conclusion}
In this paper, we reinvestigated the main limitations of Translation-based embedding models from two aspects: \textit{score} and \textit{loss}. We showed that existing theories corresponding to the limitations of the models are inaccurate because the effect of loss functions has been ignored. Accordingly, we presented new theories about the limitations by consideration of the effect of score and loss functions. We proposed TransComplEx, a new variant of TransE which is proven to be less limited comparing to the TransE. The model is trained by using various loss functions on standard dataset including FB15K, FB15K-237, WN18 and WN18RR. According to the experiments, TransComplEx with proper loss function significantly outperformed translation-based embedding models. Moreover, TransComplEx got competitive performance comparing to the state-of-the-art embedding models while it is more efficient in time and memory. The experimental results conformed the presented theories corresponding to the limitations.  

% \bibliography{emnlp-ijcnlp-2019}
% \bibliographystyle{acl_natbib}
\clearpage
\bibliography{main.bbl}
\bibliographystyle{acl_natbib}

\appendix
\clearpage

\newpage

\section{Supplementary Material}

The proof of lemmas are provided as follows:

\textit{Lemma 1:} Let assumption (a) holds, then TransE and TransComplEx cannot infer a reflexive relation pattern with non-zero relation vector. 
With assumptions (b), (c) and (d), however, this is not true anymore and the models can infer reflexive relation patterns  

\textit{Proof}
1) Let $r$ be a reflexive relation and condition a) holds. For TransE, we have 
\begin{equation}
    \begin{aligned}
    \textbf{h} + \textbf{r} - \textbf{h} = \textbf{0}.
    \end{aligned}
\end{equation}

Therefore, the relation vector collapses to a null vector ($\textbf{r} = \textbf{0}).$ As a consequence of $\textbf{r} = \textbf{0},$ embedding vectors of head and tail entities will be same which is undesired. Therefore, TransE cannot infer reflexive relation with $\textbf{r} \neq \textbf{0}.$

For TransComplEx, we have
\begin{equation}
    \begin{aligned}
    \textbf{h} + \textbf{r} - \bar{\textbf{h}} = \textbf{0}.
    \end{aligned}
\end{equation}

We have

\begin{equation}
    \begin{aligned}
        Re(\textbf{r}) = \textbf{0}, \\
        Im(\textbf{r}) = -2 Im(\textbf{h}).
    \end{aligned}
\end{equation}

Therefore, all entities will have same embedding vectors which is undesired. 

2) Using condition (b), we have

$$\| \textbf{h} + \textbf{r} - \textbf{t}\| = \gamma_1.$$
It gives $\| \textbf{r} \| = \gamma_1.$ Therefore, in order to infer reflexive relation, the length of the relation vector should be $\gamma_1.$ Consequently, TransE and TransComplEx can infer reflexive relation. The same procedure can be used for the conditions (c) and (d).

\textit{Lemma 2:} 1) Let the assumption b) or c) or d) holds. TransComplEx can infer a relation pattern which is neither reflexive nor irreflexive. 2) TransE cannot infer the relation pattern. 

\textit{proof:} 
1) Let the relation $r$ be neither reflexive nor irreflexive and two triples $(e_1,r,e_1), (e_2,r,e_2)$ be positive and negative respectively. 
Therefore the following inequalities hold:

\begin{align}
\begin{cases}
    \| \textbf{e}_1 + \textbf{r} - \Bar{\textbf{e}_1} \| \leq \lambda_1, \\
    \| \textbf{e}_2 + \textbf{r} - \Bar{\textbf{e}_2} \| \geq \lambda_2.
\end{cases}
\label{R2}
\end{align}

Equation \ref{R2} is rewritten as follows:

\begin{equation}
    \begin{aligned}
    \| Re(\textbf{r}) + i(Im(\textbf{r}) + 2 Im(\textbf{e}_1))\| \leq \gamma_1, \\
    \| Re(\textbf{r}) + i(Im(\textbf{r}) + 2 Im(\textbf{e}_2))\| \geq \gamma_2, \\    
    \end{aligned}
\end{equation}

For TransE in real space,  $\| Re(\textbf{r})\| \leq \gamma_1$ and $\| Re(\textbf{r}) \| \geq \gamma_2$ cannot be held simultaneously when $\gamma_2 > \gamma_1$. Therefore, TransE in real space cannot encode a relation  which is neither reflexive nor irreflexive. In contrast, TransE in complex space can encode the relation by proper assignment of imaginary parts of entities. Therefore, theoretically TransComplEx can infer a relation which is neither reflexive nor irreflexive. 

\textit{Lemma 3}: 1) TransComplEx can infer symmetric patterns with condition a), b), c) and d). 2) TransE cannot infer symmetric patterns with condition a) with non-zero vector for relation. 3) TransE can infer a relation pattern which is symmetric and reflexive with conditions b), c) and d). 

\textit{Proof:} 1), 2) Let $r$ be a symmetric relation and a) holds. We have

\begin{equation}
    \begin{aligned}
    \textbf{h} + \textbf{r} = \bar{\textbf{t}},\\ 
    \textbf{t} + \textbf{r} = \bar{\textbf{h}}.
    \end{aligned}
\end{equation}

Trivially, we have

\begin{equation}
    \begin{aligned}
    Re(\textbf{h}) + Re(\textbf{r}) = Re(\textbf{t}), \\
    Re(\textbf{t}) + Re(\textbf{r}) = Re(\textbf{h}), \\
    Im(\textbf{h}) + Im(\textbf{r}) = -Im(\textbf{t}), \\
    Im(\textbf{t}) + Im(\textbf{r}) = -Im(\textbf{h}), \\
    \end{aligned}
\end{equation}

For TransE in real space, there is
\begin{align*}
    Re(\textbf{h}) + Re(\textbf{r}) = Re(\textbf{t}), \\
    Re(\textbf{t}) + Re(\textbf{r}) = Re(\textbf{h}), 
\end{align*}
Therefore, $Re(\textbf{r}) = \textbf{0}.$ It means that TransE cannot infer symmetric relations with condition a). For TransComplEx, additionally we have
\begin{align*}
    Im(\textbf{h}) + Im(\textbf{r}) = -Im(\textbf{t}), \\
    Im(\textbf{t}) + Im(\textbf{r}) = -Im(\textbf{h}), 
\end{align*}
It concludes $Im(\textbf{h}) + Im(\textbf{r}) + Im(\textbf{t}) = \textbf{0}$. Therefore, TransE in complex space with condition a) can infer symmetric relation. 
Because a) is an special case of b) and c), TransComplEx can infer symmetric relations in all conditions. 

3) For TransE with condition b), there is 
\begin{align}
    \| \textbf{h} + \textbf{r} - \textbf{t} \| = \gamma_1, \\
    \| \textbf{t} + \textbf{r} - \textbf{h} \| = \gamma_1. 
    \label{910}
\end{align}
The necessity condition for encoding symmetric relation is
$\| \textbf{h} + \textbf{r} - \textbf{t} \| = \| \textbf{t} + \textbf{r} - \textbf{h} \|.$ This implies 
$
    \|h \| cos(\theta_{h,r}) = \|t \| cos(\theta_{t,r}).
$
Let $h-t = u$, by \ref{910} we have 
$
    \| \textbf{u} + \textbf{r}  \| = \gamma_1, 
    \| \textbf{u} - \textbf{r} \| = \gamma_1. 
    \label{15}
$

Let $\gamma_1 = \alpha \|r\|$. We have
\begin{align}
\begin{cases}
    \|\textbf{u}\|^2 + (1-\alpha^2) \|\textbf{r}\|^2 = - 2 \langle \textbf{u},\textbf{r}\rangle \\ 
    \|\textbf{u}\|^2 + (1-\alpha^2) \|\textbf{r}\|^2 =  2 \langle \textbf{u},\textbf{r}\rangle
\end{cases}
\label{16}
\end{align}
Regarding \ref{16}, we have

$\|\textbf{u}\|^2 + (1-\alpha^2) \|\textbf{r}\|^2 = -(|\textbf{u}\|^2 + (1-\alpha^2) \|\textbf{r}\|^2) $.

$ \rightarrow \|\textbf{u}\|^2 = (\alpha^2 - 1) \|\textbf{r}\|^2.$

To avoid contradiction,  $\alpha \geq 1$. If $\alpha \geq 1$ we have $ \cos(\theta_{u,r}) = \pi /2 .$
Therefore, TransE can encode symmetric pattern with condition b), if $\gamma_1 = \alpha \|r\|$ and  $\alpha \geq 1$. From the proof of condition b), we conclude that TransE can encode symmetric patterns under conditions c) and d).

\textit{Lemma 4:} 1) Let a) holds. Limitation L4 holds for both TransE and TransComplEx. 2) Limitation L4 is not valid when assumptions b), c) and d) hold. 

\textit{Proof:} 
1) The proof of the lemma with condition a) for TransE is mentioned in the paper \cite{kazemi2018simple}. For TransComplEx, the proof is trivial. 
2) Now, we prove that the limitation L4 is not valid when b) holds. 

Let condition b) holds and relation $r$ be reflexive, we have
$    \| \textbf{e}_1 + \textbf{r} -\textbf{e}_1\| = \gamma_1,
    \| \textbf{e}_2 + \textbf{r} -\textbf{e}_2\| =\gamma_1.
$

Let $\| \textbf{e}_1 + \textbf{r} - \textbf{e}_2 \| = \gamma_1$. To violate the limitation L4, the triple $(e_2, r, e_1)$ should be negative i.e.,

$\| \textbf{e}_2 + \textbf{r} - \textbf{e}_1 \| > \gamma_1,$

$\rightarrow \| \textbf{e}_2 + \textbf{r} - \textbf{e}_1 \|^2 > \gamma_1^2,$

$\rightarrow \| \textbf{e}_2 \|^2 + \| \textbf{e}_1 \|^2 + \| \textbf{r} \|^2 + 2 <\textbf{e}_2,\textbf{r}> - 2 <\textbf{e}_2,\textbf{e}_1> - 2 <\textbf{e}_1,\textbf{r}> \,\, > \gamma_1^2.$

Considering $\| \textbf{e}_1 + \textbf{r} - \textbf{e}_2 \| = \gamma_1$, we have

$
     <\textbf{e}_2,\textbf{r}> - <\textbf{e}_1,\textbf{r}> \,\, > 0, \\
     \rightarrow \, <\textbf{e}_2 - \textbf{e}_1,\textbf{r}> \,\, > 0, \\
     \rightarrow \, cos(\theta_{(\textbf{e}_2 - \textbf{e}_1), \textbf{r}}) \,\, > 0,
$

Therefore, the limitation L4 is not valid i.e., if a relation $r$ is reflexive, it may not be symmetric.
TransE is special case of TransComplEx and also condition b) is special case of condition c). Therefore using conditions b), c) and d), the limitation L4 is not valid for TransE and TransComplEx.

\textit{Lemma 5:}  1) Under condition a), the limitation L5 holds for both TransE and TransComplEx. 2) Under conditions b), c) and d), L5 is not valid for both TransE and TransComplEx. 

\textit{proof}

1) Under condition a), equation $\textbf{h} + \textbf{r} - \textbf{t} = \textbf{0}$ holds. Therefore, according to the paper \cite{kazemi2018simple}, the model has the limitation L5. 

2) If a relation is reflexive, with condition b), we have
$
    \| \textbf{e}_1 + \textbf{r} -\textbf{e}_1\| = \gamma_1,
    \| \textbf{e}_2 + \textbf{r} -\textbf{e}_2\| =\gamma_1.
$
Therefore, $\|r\|= \lambda_1.$
Let 
\begin{align}
    \begin{cases}
    \| \textbf{e}_1 + \textbf{r} -\textbf{e}_2\| = \gamma_1,\\
    \| \textbf{e}_2 + \textbf{r} -\textbf{e}_3\| =\gamma_1.
    \end{cases}
\label{22}
\end{align}

we need to show the following inequality wouldn't give contradiction:
$
    \| \textbf{e}_2 + \textbf{r} -\textbf{e}_3\|  >
\gamma_1.
$

From \ref{22} we have
$< \textbf{e}_2, (\textbf{e}_1 + \textbf{e}_2 + \textbf{e}_3) >  < 0,$ which is not contradiction. 

Therefore, with conditions b) and c), the limitation L5 is not valid for both TransE and TransComplEx. 

\textbf{Limitation L6:}
\textit{Lemma 6:} 1) With condition (a), the limitation L6 holds for both TransE and TransComplEx. 2) With conditions (b), (c) and (d), the limitation L6 doesn't hold for the models. 

\textit{Proof}: 
1) With condition (a), the limitation L6 is proved in \cite{kazemi2018simple}.
2) Considering the assumption of L6 and the condition (b), we have

\begin{align}
    \begin{cases}
    \| \textbf{e}_1 + \textbf{r} -\textbf{s}_1\| = \gamma_1,\\
    \| \textbf{e}_1 + \textbf{r} -\textbf{s}_2\| =\gamma_1.\\
    \| \textbf{e}_2 + \textbf{r} -\textbf{s}_1\| =\gamma_1.
    \end{cases}
    \label{24}
\end{align}

We show the condition that $\| \textbf{e}_2 + \textbf{r} -\textbf{s}_2\| > \gamma_1$ holds.

Substituting \ref{24} in  $\| \textbf{e}_2 + \textbf{r} -\textbf{s}_2\| > \gamma_1$, we have 

$cos(\theta_{(s_1-s_2),(e_1-e_2)}) < 0.$ Therefore, there are assignments to embeddings of entities that the limitation L6 is not valid with condition (b), (c) and (d). 

\begin{figure}[h!]
\centering
\includegraphics[width=0.45\textwidth]{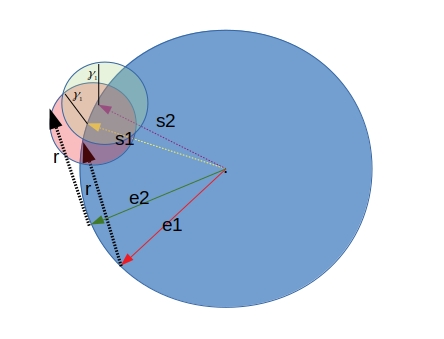} %need a better resolution: width=1.0\textwidth
\caption{Investigation of L6 with condition (c): The limitation is not valid, because the triple ($e_2,r,s_2$) can get an score to be considered as negative while triples ($(e_1,r,s_1), (e_1,r,s_2), (e_2,r,s_1)$) are positive.}
	\label{fig:l6}
\end{figure}

Figure \ref{fig:l6} shows that the limitation L6 is invalid by proper selection of loss function. 

\subsection{Further limitations and future work}

In the paper, we have investigated the six limitations of TransE which are resolved by revision of loss function. However, revision of loss functions can resolve further limitations including 1-N, N-1 and M-N relations. More concretely, setting upper-bound for the scores of positive samples can mitigate the M-N problem. We will leave it as future work. 

Our theories can be extended to every distance-based embedding models including RotatE etc. 

Moreover, the negative likelihood loss has been shown to be effective for training different embedding models including RotatE and TransE. This can also be explained by reformulation of negative likelihood loss as standard optimization problem, showing the the loss put a boundary for the score functions.

We will consider the mentioned points as future work.

\end{document}